\documentclass[conference]{IEEEtran}
\IEEEoverridecommandlockouts

\usepackage{hyperref}
\usepackage{cite}
\usepackage{float}
\usepackage{amsmath,amssymb,amsfonts}
\usepackage{algorithmic}
\usepackage{graphicx}
\usepackage{textcomp}
\usepackage{xcolor}
\usepackage{pdfpages}
\usepackage{pgfplots}
\usepackage{booktabs}
\usepackage{multirow}
\usepackage{caption}
\usepackage{subcaption}
\usepackage{setspace}
\usepackage{multicol}
\usepackage{rotating}
\usepackage{makecell}

\pgfplotsset{compat = 1.3}

\makeatletter
\newcommand{\removelatexerror}{\let\@latex@error\@gobble}
\makeatother

\usepackage[norelsize,ruled,vlined,linesnumbered]{algorithm2e}
\DontPrintSemicolon

\SetKwSty{texttt}
\SetKw{With}{with}
\SetKw{And}{and}
\SetKw{Continue}{continue}
\SetKwRepeat{Do}{do}{while}
\SetKwFor{With}{with}{do}{end}
\SetKwFor{ParallelForEach}{foreach}{in parallel}{end}
\SetKwProg{fn}{}{}{end}

\pgfplotsset{
  log x ticks with fixed point/.style={
    xticklabel={
      \pgfkeys{/pgf/fpu=true}
      \pgfmathparse{exp(\tick)}%
      \pgfmathprintnumber[fixed relative, precision=3]{\pgfmathresult}
      \pgfkeys{/pgf/fpu=false}
    }
  },
  log y ticks with fixed point/.style={
    yticklabel={
      \pgfkeys{/pgf/fpu=true}
      \pgfmathparse{exp(\tick)}%
      \pgfmathprintnumber[fixed relative, precision=3]{\pgfmathresult}
      \pgfkeys{/pgf/fpu=false}
    }
  }
}

\let\oldnl\nl% Store \nl in \oldnl
\newcommand{\nonl}{\renewcommand{\nl}{\let\nl\oldnl}}% Remove line number for one line

\SetCommentSty{emph}

\newcommand{\todo}[1]{}
\renewcommand{\todo}[1]{{\color{red} TODO: {#1}}}

\def\BibTeX{{\rm B\kern-.05em{\sc i\kern-.025em b}\kern-.08em
    T\kern-.1667em\lower.7ex\hbox{E}\kern-.125emX}}
\begin{document}

% \title{Efficient Dense Map Representations for Mobile Robotics using VDB Datastructure}
\title{Efficient Global Occupancy Mapping for Mobile Robots using OpenVDB
  \thanks{This work has been conducted within the competence center ROBDEKON – Robotic Systems for Decontamination in Hazardous Environments, which is funded by the Federal Ministry of Education and Research (BMBF) within the scope of the German Federal Government’s Research for Civil Security program under grant no. 13N14674.}
}

\makeatletter
\newcommand{\linebreakand}{%
\end{@IEEEauthorhalign}
\hfill\mbox{}\par
\mbox{}\hfill\begin{@IEEEauthorhalign}
}
\makeatother

\author{
\IEEEauthorblockN{1\textsuperscript{st} Raphael Hagmanns}
\IEEEauthorblockA{
\textit{Karlsruhe Institute for Technology (KIT)} and\\
\textit{Fraunhofer Institute of Optronics,}\\
\textit{System Technologies and Image Exploitation}\\
Karlsruhe, Germany\\
raphael.hagmanns@kit.edu
}
\and
\IEEEauthorblockN{2\textsuperscript{nd} Thomas Emter}
\IEEEauthorblockA{
  \textit{Fraunhofer Institute of Optronics,}\\
  \textit{System Technologies and Image Exploitation}\\
  Karlsruhe, Germany\\
  thomas.emter@iosb.fraunhofer.de\\[5mm]
}
\linebreakand
\IEEEauthorblockN{3\textsuperscript{rd} Marvin Grosse Besselmann}
\IEEEauthorblockA{
  \textit{FZI Forschungszentrum Informatik}\\
  Karlsruhe, Germany\\
  grossebesselmann@fzi.de
}
\and
\IEEEauthorblockN{4\textsuperscript{th} Jürgen Beyerer}
\IEEEauthorblockA{
  \textit{Karlsruhe Institute for Technology (KIT)} and\\
  \textit{Fraunhofer Institute of Optronics,}\\
  \textit{System Technologies and Image Exploitation}\\
  Karlsruhe, Germany \\
}
}

\maketitle
\begin{abstract}
In this work we present a fast occupancy map building approach based on the VDB datastructure. Existing log-odds based occupancy mapping systems are often not able to keep up with the high point densities and framerates of modern sensors. Therefore, we suggest a highly optimized approach based on a modern datastructure coming from a computer graphic background. A multithreaded insertion scheme allows occupancy map building at unprecedented speed. Multiple optimizations allow for a customizable tradeoff between runtime and map quality. We first demonstrate the effectiveness of the approach quantitatively on a set of ablation studies and typical benchmark sets, before we practically demonstrate the system using a legged robot and a UAV.
\end{abstract}


\section{Introduction}\label{sec:intro}

A detailed understanding of a potentially unknown environment plays a fundamental role in mobile robotic applications. Different robots and environments come along with varying requirements for the map building process in terms of accuracy, efficiency and usablility. Common SLAM methods, which attempt to map an environment while simultaneously localizating the robot in it, usually have to find a balance between these properties. Due to the complexity of the task, it is still very challenging to perform SLAM while maintaining a dense map representation. Compared to commonly used sparse map representation, \textit{occupancy grids} have many advantages as they integrate all available information into a single representation which is easy to understand for an operator and also allows for efficient queries.

2D projections of such dense reprenstations have been used extensively for mobile robot navigation tasks. As robots become more agile, scenes more complex and sensors more capable, it is also desireable to adopt these structures for the third dimension. For ground vehicles, so called 2.5D elevation maps as suggested by Herbert et.~al~\cite{herbertmapping} may be sufficient. However, for agile robots with complex kinematics such as legged robots or UAVs, a full 3D representation of the environment is essential.

\begin{figure}[tb]
  \centering
  \includegraphics[width=0.7\columnwidth]{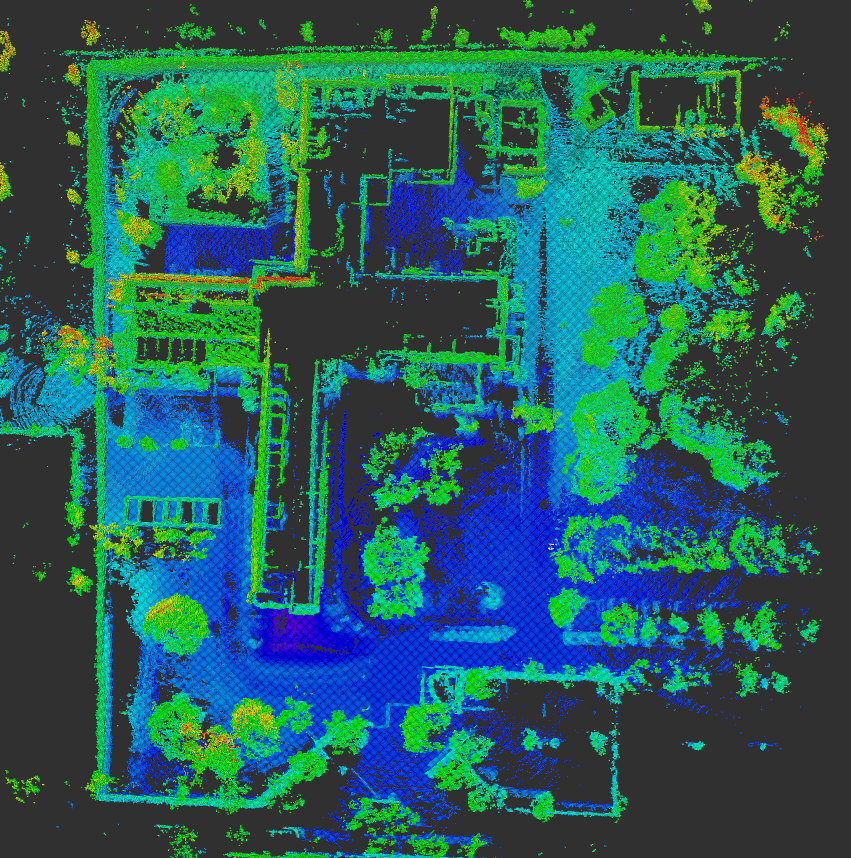}
  \caption{Map of the Fraunhofer IOSB site using the mapping approach. Preprocessing using a custom factor graph approach described in \cite{iosb_mapping}.}
  \label{fig:mapping_iosb}
  \vspace{-0.5cm}
\end{figure}

To store the 3D map memory efficient and with certain complexity guarantees for random access operations, different datastructures have been suggested. The most prominent example for such a 3D map representaion is the OctoMap~\cite{octomap} framework by Hornung et al., working on octrees as hierarchical tree structure. It allows for a memory effiecient storing through efficient pruning and propagating leaf states to higher levels of the tree. OctoMap has been considered state-of-the-art for a long time but as sensors are achieving higher rates with millions of points per second, the \textit{insert} operation of OctoMap is not able to achieve real-time performance. As the octree has a fixed layout, it is also difficult to later increase the volume without performance overhead. One of the most popular frameworks for global consistent mapping is VoxBlox by Oleynikova et al.~\cite{voxblox}. They incrementally build a \textit{Truncated Signed Distance Map} (TSDF)~\cite{sdf} instead of an occupancy voxel grid and reach almost real-time performance on a single core implementation using a hashmap representation instead of a tree as fundamental datastructure.

OpenVDB is a modern framework developed in a computer graphic context. Similar to OctoMap it is based on a hieracical structure but it comes with certain accelerators to support almost constant insertion and read procedures using a B+ tree like structure. The main reason for the improved performance is an advanced indexing and caching system. We therefore leverage OpenVDB as underlying datastructure for our map building approach and present its capabilities in further detail in the upcoming Section~\ref{sec:pipeline}. OpenVDB as flexible backbone allows not only fast insertions but also supports efficient raycasting step samplers and a virtually infinite map size.

Only few works utilize the VDB datastructure as occupancy representation so far. Our work was originally based on~\cite{vdb_fzi} by Besselmann et al. and can be considered a successor with a revised update scheme and optimized insertion procedure. They bring up the idea to integrate data into a temporary grid first to cope with discretization ambiguities which arise when raycasting new data.  In~\cite{vdb-edt}, Zhu et al. present a full framework for occupancy and distance mapping, which also uses a raycast based insertion scheme in order to create the occupancy map. They put the focus on the \textit{Euclidean Distance Transform} step and disregard the map integration itsself. Macenski et~al.~\cite{spatiotemporal} built a spatio-temporal voxel layer on top of OpenVDB. They focus on local dynamic maps and therefore use a sensor frustrum based visibility check instead of raycasting as integration scheme.

In our work we further push the limits of the underlying OpenVDB structure by supporting a flexible multithreaded raycasting insertion scheme into the map supported by additional ray-level hashing to avoid unnecessary operations. Fast merging operations of single \textit{bit-grids} allow for a minimal lock time for modifying the global occupancy grid. Different subsample strategies can be selected to allow for a dynamically adjustable tradeoff between map accuracy and efficiency.

The main contribution of the work can be summarized as follows:
\begin{itemize}
  \item We present a real-time capable and multithreaded dense mapping approach for efficiently creating occupancy maps based on the VDB data structure.
  \item We introduce several optimizations in the integration scheme allowing for a user-definable tradeoff between map accuracy and efficiency.
  \item We conduct benchmarks on different operations and test the whole pipeline in simulation and real environments.
  \item We open-source the codebase and a corresponding wrapper for the Robot Operating System II (ROS II~\cite{ros2}) which enables fast prototyping for mobile robotic applications.
\end{itemize}

The remainder of this work is structured as follows. Section~\ref{sec:pipeline} formalizes the problem of mapping and introduces OpenVDB as underlying datastructure. We also give a detailed overview on the insertion scheme and introduce various optimizations leading to improved performance. In Section~\ref{sec:evaluation} we carry out different experiments to verify the effect of the introduced optimizations. We summarize and conclude the work and discuss potential future improvements in Section~\ref{sec:discussion}.

\section{Mapping Pipeline}\label{sec:pipeline}

In this section, we first formalize the problem before discussing the proposed insertion scheme.

\subsection{Problem Statement}

The main goal of occupancy mapping is to create a map \(\mathcal{M}_{\text{occ}}\) which stores the occupancy probability \(p(x_{i}|s_{1:t}, z_{1:t})\) for each cell \(x_{i} \in \mathcal{M}_{\text{occ}}\) given some sensor measurements \(s_{1:t}\) and the corresponding robot poses \(z_{1:t}\), where \(1{:}t\) denotes the sequence from the start up to time \(t\). We consider a cell to be an obstacle if \(p(x_{i}|s_{1:t},z_{1:t})\) exceeds a certain threshold \( \phi_{\text{occ}}\) and free if it falls below the threshold \(\phi_{\text{free}}\). Note that being marked as \textit{free} is different from not being observed yet.

\subsection{OpenVDB}

\begin{figure}[b]
  \centering
  \includegraphics[width=1\columnwidth]{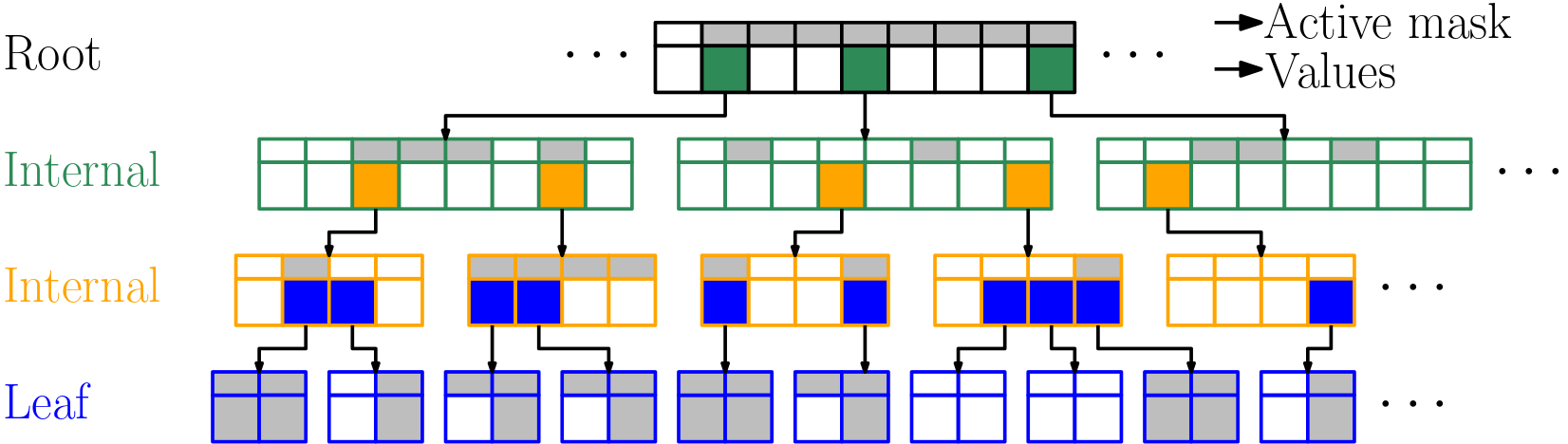}
  \caption{Illustration of the underlying VDB datastructure. Image is adopted from the original publication \cite{vdb}. The height of the tree is typically 4 with one root node (gray) and two internal layers (green, orange) as well as leaf nodes, which store the actual tile values (blue). All nodes in lower levels have a branching factor equal to a power of two. Root and internal nodes store pointers to their respective child nodes. An \textit{active} bitmask for each nodes encodes if subsequent tiles are active or not (gray values).}
  \label{fig:vdb}
\end{figure}

The OpenVDB framework has been introduced by Museth et~al. in 2013 \cite{vdb}. Originally it was designed for computer graphic applications such as rendering animations of complex mesh structures and time-varying sparse volumes such as clouds. Since then, it has been widely adopted to different applications as it allows for flexible modifications to its core structure. At its heart it levereges a B+ tree~\cite{bplustree} variant as main datastructure. This structure is supported by hieracically organized caches to faciliate fast access to inner tree nodes. Such a datastructure is ideal to store sparse voxelized environment representations. The discretization of the space can be adjusted by choosing the size of the leaf nodes accordingly. Other adjustable parameters are the tree depth and branching factors which can further improve the memory footprint depending on the sparsity of the environment. Typical branching factors for the VDB datastructures are very large compared to the octree branching factors, which are typically two in each spatial dimension and thus lead to less shallow trees.
The schematics of the underlying tree structure is depicted in Figure~\ref{fig:vdb}. The fixed height of the tree makes it possible to implement \textit{insert} and \textit{read} operations in constant time on average~\cite{vdb}. It also allows the framework to efficiently utilize typical cache architectures on modern CPUs to further speed up read operations for tiles with spatial proximity. Access to the tree's tile values is implemented via a virtually infinite \textit{index space}, which can be accessed by signed \(32\)-bit integer coordinates, allowing the map to grow in each direction without additional overhead. As noted in Figure~\ref{fig:vdb}, tree nodes additionally store bitsets indicating if subsequent nodes are active or not. This allows for a fast traversal of a sparse volume without the need to visit tile values explicitly. The features of the tree structure are described in further detail in the original work~\cite{vdb}. Moreover, VDB allows for efficient raycasting operations using optimized index space iterators. Therefore, we utilize a raycast based sensor insertion scheme which is described in further detail in the following section.

\subsection{Map Updates}
The proposed structure is not tied to a specific sensor. Typical data comes from a LiDAR sensor, mimicing the raycast operation which is also performed to create the obstacle map. Cells store their occupancy probability in order to reflect not only occupied but also free space. To represent time-dependent updates we faciliate the very commonly used log-odds based update scheme initially formulated by Moravec and Elfes in~\cite{logodds}. Essentially we calculate
\begin{equation}
  \mathcal{M}_{\text{occ}}(x_{i}|s_{1:t})= \mathcal{M}_{\text{occ}}(x_{i}|s_{1:t-1}) + \log\left[\frac{p(x_{i}|s_{1:t})}{1-p(x_{i}|s_{1:t})}\right]
  \label{eq:update}
\end{equation}
in each update step and for each \(x_{i} \in \mathcal{M}_{\text{occ}} \).

Algorithm~\ref{alg:map} gives an abstracted overview on the insertion process. Essentially we do parallel raycasting in coaligned temporary grids, which are merged together in a later step. We first create a coaligned map \(\mathcal{M}_{\text{agg}}\) (line 2), which we later use to aggregate the temporary maps. We then divide the incoming points into different chunks which can be processed in parallel (line 5). Points of each chunk are inserted by calculating their respective end position in world coordinates (line 9) and marching along the ray using OpenVDBs digital differential analyzer (DDA) implementation (lines 19-22) and marking all visited voxels as \textit{active}. The only place where we need to lock the threads is during the merge operation in lines 23-24. This can be done efficiently as we simply XOR the boolean grids together.
As OpenVDB stores the active state of nodes in a fast accessible bitmask, we can now efficiently iterate over all \textit{active} values in our aggregated map \(\mathcal{M}_{\text{agg}}\) (line 25). We increase or decrease the occupancy value following Equation~\ref{eq:update} (lines 27 to 29). If a voxel exceeds or falls below a certain threshold it will be marked as occupied or unoccupied.

We suggest and implement two runtime optimizations, which are roughly based on similar ideas used in Voxblox~\cite{voxblox}, namely a \textit{subsampling} and a \textit{bundling} optimiztation. \textit{Subsampling} (cf.  Figure~\ref{fig:subsample}) uses an additional map \(\mathcal{M}_{\text{sub}}\) which increases the resolution \(\mathcal{M}_{\text{occ}}\) by a \textit{subsampling-factor} \(\delta_{\text{sub}}\). Typically, \(\delta_{\text{sub}}\) is restricted to powers of \(2\), we use \(4\) in most setups. In addition we use a Hashmap \(\mathcal{H}_{\text{sub}}\) (line 3) storing for each cell in \(\mathcal{M}_{\text{sub}}\), if it has already been visited in the current integration step. If this is the case, all subsequent integrations are skipped (lines 12 and 13). This optimization comes with the cost of inaccurate details but can save a lot of integration steps especially in dense environments with large voxel sizes, where a single voxel is hit multiple times.

\removelatexerror
\begin{algorithm}[H]
  \caption{Map Update Scheme}
  \label{alg:map}
  \setstretch{1.30}
  \SetKwFunction{MapUpdate}{MapUpdate}
  \KwIn{Pointcloud $\mathcal{P}$, Sensor origin $\mathbf{o}$, Number of Chunks $c$, Global Occupancy Map $\mathcal{M}_{\text{occ}}$}
  $\mathcal{M}_{\text{sub}} \leftarrow \text{coalign}(\mathcal{M})$ \tcp{Aligned Subsampling Map}
  $\mathcal{M}_{\text{agg}} \leftarrow \text{coalign}(\mathcal{M})$ \tcp{Aligned Aggregation Map}
  Hashmap$\left<\text{coord } x,\text{bool hit}\right> \mathcal{H}_{\text{sub}} \leftarrow [\;]$\;
  Hashmap$\left<\text{int count}, \text{coord } x, \text{bool maxray}\right> \mathcal{H}_{\text{bun}} \leftarrow [\;]$ \;
  $\mathcal{P}_{i} \leftarrow \text{Equal Chunks of } \mathcal{P} \text{ for } i=1..c$\;
  \ParallelForEach{$\mathcal{P}_{i} \in \mathcal{P}$}{
    $\mathcal{M}_{\text{temp}} \leftarrow \text{coalign}(\mathcal{M})$ \tcp{Aligned Temporary Map}
    \ForEach{$\mathbf{p} \in \mathcal{P}_{i}$} {
      $\mathbf{r}_{\text{end}}=\mathbf{o} + (\mathbf{p} - \mathbf{o}) \text{ in } \mathcal{M}_{\text{agg}}$ \;
      $\mathbf{r}_{\text{end\_sub}}=\mathbf{o} + (\mathbf{p} - \mathbf{o}) \text{ in } \mathcal{M}_{\text{sub}}$ \;
      is\_maxray $\leftarrow $ check for max-length ray \;
     \If{$\mathcal{H}_{\text{sub}}[\mathbf{r}_{\text{end\_sub}}]$}{
        \Continue\;
      }
      \eIf{$\mathcal{H}_{\text{bun}}[\mathbf{r}_{\text{end}}].\text{count}>\text{thresh}$}{
        $(\text{count}, \overline{\mathbf{r}_{\text{end}}}, \text{maxray}) = \mathcal{H}_{\text{bun}}[\mathbf{r}_{\text{end}}]$ \;
      }{
        $\mathcal{H}_{\text{bun}}[\mathbf{r}_{\text{end}}]+=(1,\mathbf{r}_{\text{dir}},\text{is\_maxray})$ \;
        \Continue\;
      }
      \Do{$\mathbf{r}_{\text{dda}} \neq \mathbf{r}_{\text{end}}$}{
        $\mathcal{M}_{\text{temp}}[\mathbf{r}_{\text{dda}}].\text{active} = \text{true}$\;
        $\mathbf{r}_{\text{dda}}++$
      }

    }
    \With{MapLock$(\mathcal{M}_{\text{temp}})$}{
      $\mathcal{M}_{\text{agg}} \;|=\; \mathcal{M}_{\text{temp}}$ \;
    }
    }
  \ForEach{active value $x \in \mathcal{M}_{\text{agg}}$}{
    \eIf{$x$}{
      increase occupancy on $\mathcal{M}_{\text{occ}}[x]$ following Eq.~\ref{eq:update}\;
      % \If{$\mathcal{M}_{\text{obs}}[x] > \phi_{\text{occ}}$}
      % {
      %   $\textsc{setObstacle}(\mathcal{M}_{\text{dist}}[x])$
      % }

    }{
      reduce occupancy on $\mathcal{M}_{\text{occ}}[x]$ following Eq.~\ref{eq:update}\;
      % \If{$\mathcal{M}_{\text{obs}}[x] < \phi_{\text{free}}$}
      % {
      %   $\textsc{removeObstacle}(\mathcal{M}_{\text{dist}}[x])$
      % }
    }
  }
  \KwOut{Updated $\mathcal{M}_{\text{occ}}$}
  % \vspace*{.4cm}
\end{algorithm}

\begin{figure*}[tb]
  \centering
  \begin{subfigure}[b]{0.35\textwidth}
    \captionsetup{margin={2.6cm,0cm}}
    \centering
    \includegraphics[page=2, width=\textwidth]{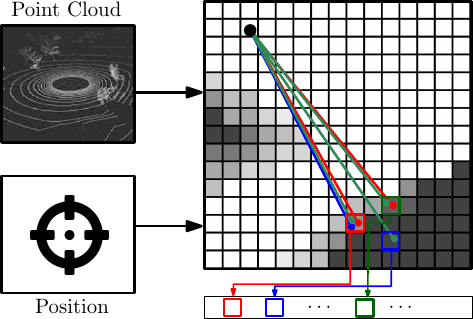}
    \caption{Default Raycasting}
    \label{fig:raycasting}
  \end{subfigure}
  \hfill
  \begin{subfigure}[b]{0.2\textwidth}
    \centering
    \includegraphics[page=5, width=\textwidth]{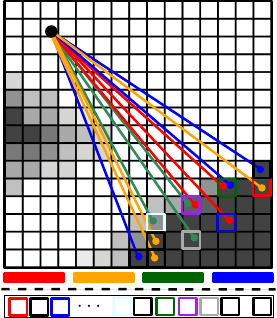}
    \caption{Parallelization}
    \label{fig:parallel}
  \end{subfigure}
  \hfill
  \begin{subfigure}[b]{0.2\textwidth}
    \centering
    \includegraphics[page=4, width=\textwidth]{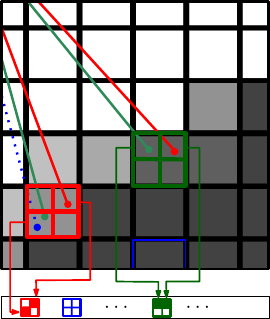}
    \caption{Subsample Optimization}
    \label{fig:subsample}
  \end{subfigure}
  \hfill
  \begin{subfigure}[b]{0.205\textwidth}
    \centering
    \includegraphics[page=3, width=\textwidth]{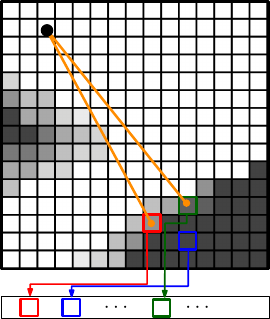}
    \caption{Bundle Optimization}
    \label{fig:bundle}
  \end{subfigure}
  \caption{Scheme of different optimizations on the insertion procedure. In (a), the standard raycasting process is visualized. When enough rays hit a voxel so that \(\phi_{\text{occ}}\) is exceeded, this voxel will be marked as occupied. Figure (b) visualizes the process of dividing rays into different chunks which can then be processed in a multithreaded fashion. In (c) the target space is further subsampled with a customizable subsampling map \(\mathcal{M}_{\text{sub}}\). Rays which hit an already marked subsampled cell (the blue ray) are skipped. In (d) multiple rays are aggregated and averaged if they hit the same target cell. }
  \label{fig:mapping_optimizations}
\end{figure*}
The \textit{bundling} optimization on the other hand bundles multiple rays together. Again, we use a hashmap \(\mathcal{H}_{\text{bun}}\) where we insert incoming ray end points without actually integrating the rays (lines 17 and 18). If a certain threshold is exceeded, we integrate the whole bundle targeting the end cell \(\mathbf{r}_{\text{end}}\) at once. In the map, we additionally store the original end points and if the ray reached its maximum length. During the integration of the bundle this information is used to average the final end point  \(\overline{\mathbf{r}_{\text{end}}}\). Again, this optimization favors environments with a lot of redundant integrations. Figure~\ref{fig:bundle} indicates that only one orange bundle is inserted into the map, even if multiple rays hit the cell (cf. Figure~\ref{fig:raycasting}). Both optimizations can be enabled or disabled individually or together. While enabling the optimizations leads to a deliberately impaired map accuracy, it can be useful as more sensor data can be integrated overall due to the additionally gained performance.
It is worth noticing that it is not necessary to deal with discretization ambiguities introduced by sequential raycasting presented in~\cite{vdb_fzi} as we use the same two step approach as in~\cite{vdb_fzi}: First we activate all visited voxels in a seperate aggregation map  \(\mathcal{M}_{\text{agg}}\) before  we integrate it into the global map \(\mathcal{M}_{\text{occ}}\).

\section{Evaluation}\label{sec:evaluation}

We will now present the results of experiments which we conducted to measure the performance of our proposed methods under different conditions. We first compare different iterations of our method in a set of ablation studies to measure the effect of different optimizations. In a next step we compare the method on typical benchmark sets before we finally conduct some real world experiments by capturing outdoor and indoor scenes of our lab. All experiments are performed using a machine equipped with a 6-core Intel\copyright Core\texttrademark i7-10850H and 32~GB of memory. As hardware platforms to carry out our experiments we use a BostonDynamics Spot equipped with an Ouster OS0-64 LiDAR for outdoor experiments as well as a custom UAV platform with a solid-state LiDAR for indoor experiments (see Figure~\ref{fig:robots} for details).

\begin{figure}[tb]
  \centering
  \begin{subfigure}[b]{0.45\columnwidth}
    \captionsetup{format=hang}
    \centering
    \includegraphics[width=\textwidth]{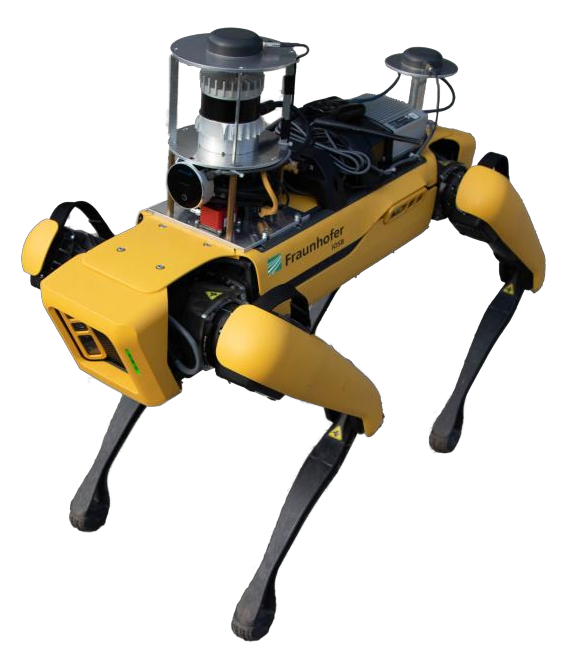}
    \caption{BostonDynamics Spot with additional sensors}
    \label{fig:spot}
  \end{subfigure}
  \hfill
  \begin{subfigure}[b]{0.45\columnwidth}
    \captionsetup{format=hang}
    \centering
    \includegraphics[width=\textwidth]{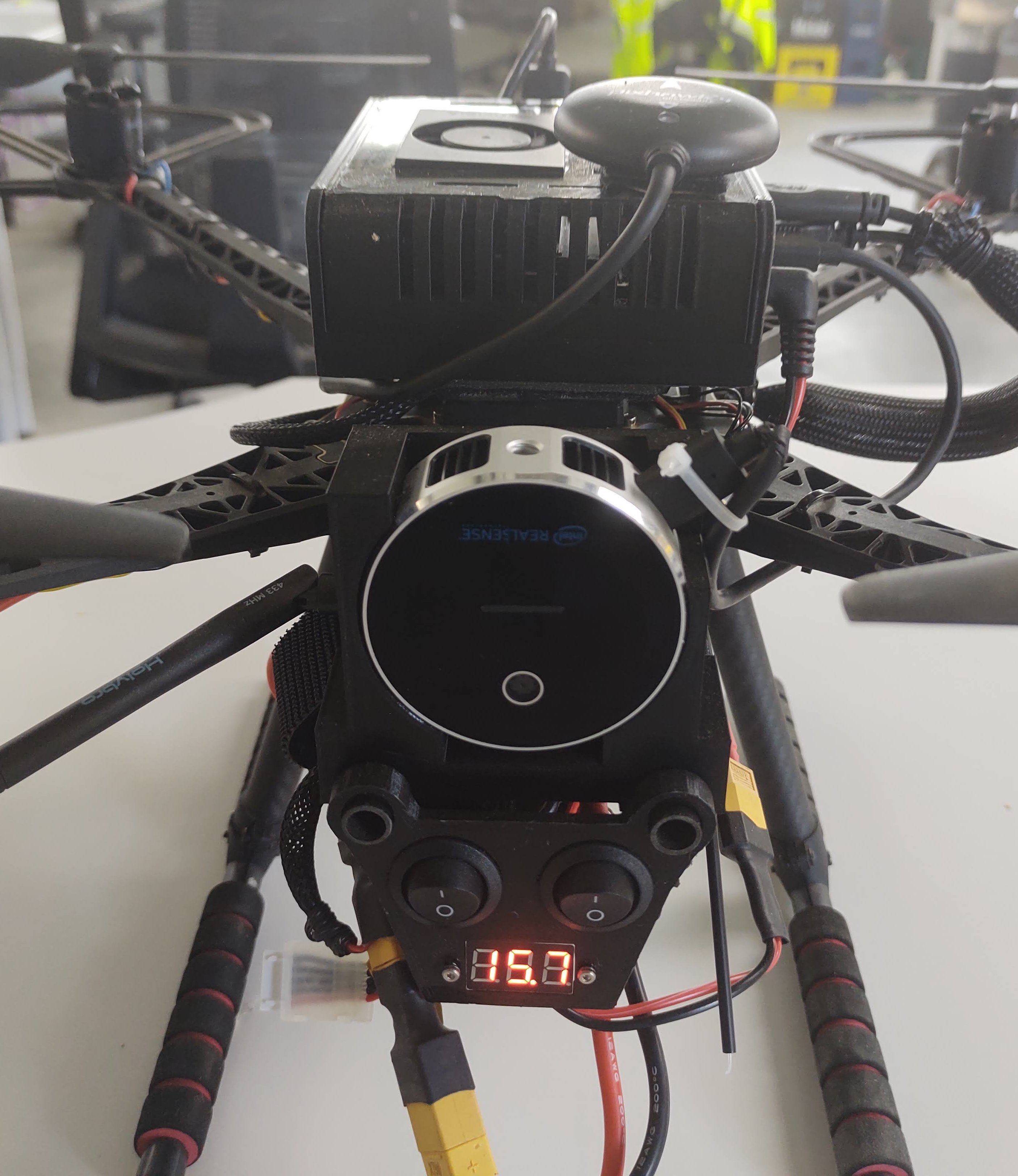}
    \caption{Custom UAV platform for indoor usage}
    \label{fig:drone}
  \end{subfigure}
  \caption{The hardware setup used for indoor and outdoor experiments. Both robots are equipped with a solid-state LiDAR (Realsense L515) and provide a localization coming from a multisensor-fusion approach.}
  \label{fig:robots}
  \vspace*{-0.3cm}
\end{figure}

\subsection{Ablation Studies}
\begin{table}[tb]
  \centering
  \caption{Runtime of different variants with 0.1m map resolution and $n$ insertions each. \textit{Random} refers to randomly sampled points in a radius which is $1.2 \cdot \text{ray length}$ as depicted in (b). The \textit{structured} environment (a) refers to a sampling where $(x,y)$ coordinates are sampled randomly but $z$ coordinates are restricted to a small width of $1$m simulating a wall structure.~(*) VDB-BUN and VDB-SUB approaches are only included as rough estimate and do not compare to the other methods, as some rays are not casted, when using bundle or subsampling optimizations. The parallel version VDB-PAR runs on 12 threads and VDB-FMAP is the improved and parallel variant restricted to a single core and with disabled optimizations while VDB-MAP is the VDB based method described in \cite{vdb_fzi}.}
  \label{tab:runtime}
  \begin{tabular}{@{}cl@{}rrrr@{}}
  \toprule
    \multirow{3}{*}{$n$} & \multirow{3}{*}{method} & \multicolumn{4}{c}{runtime [ms]} \\
            & & \multicolumn{2}{c}{ray length 6m} & \multicolumn{2}{c}{ray length 60m} \\
                          & & structure & random & structure & random \\ \midrule
  \multirow{6}{*}{\rotatebox[origin=c]{90}{50\,000}}& OctoMap~\cite{octomap} & 110 & 747 & 11676 & 42\,593 \\
                         & VDB-EDT~\cite{vdb-edt}     & 102 & 106 & 3\,005 & 3\,973\\
                            & VDB-MAP~\cite{vdb_fzi}     & 55 & 74 & $\mathbf{863}$ & $\mathbf{1\,984}$ \\
                            & VDB-FMAP      & 54 & 71 & 906 & 2\,075 \\
                            & VDB-SUB*   & 56 & 75  & 923 & 2\,149 \\
                            & VDB-BUN*      & 31 & 40 & 491 & 1\,083 \\
                            & VDB-PAR    & $\mathbf{10}$ & $\mathbf{20}$ & 1308 & 5\,581 \\ \midrule
    \multirow{6}{*}{\rotatebox[origin=c]{90}{500\,000}} & OctoMap~\cite{octomap} & 796 & 2\,189 & 44\,072 & 223\,672 \\
                         & VDB-EDT~\cite{vdb-edt}      & 957 & 987 & 22\,010 & 28\,684 \\
                          & VDB-MAP~\cite{vdb_fzi}     & 547 & 709  & 6\,042 & 11\,364 \\
                          & VDB-FMAP     & 547 & 671  & 5\,779 & 10\,718 \\
                          & VDB-SUB*  & 288 & 583  & 5\,011 & 10\,998 \\
                          & VDB-BUN*     & 261 & 330  & 2\,202 & 4\,764 \\
                          & VDB-PAR   & $\mathbf{54}$ & $\mathbf{74}$ & $\mathbf{1\,979}$ & $\mathbf{10\,339}$ \\ \bottomrule \\
  \end{tabular}
  \begin{subfigure}[b]{0.23\textwidth}
    \includegraphics[width=\textwidth]{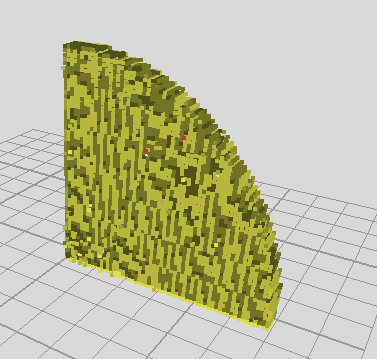}
    \caption{Structured Raycasting}
    \label{fig:structure_raycast}
  \end{subfigure}
  \hfill
  \begin{subfigure}[b]{0.23\textwidth}
    \includegraphics[width=\textwidth]{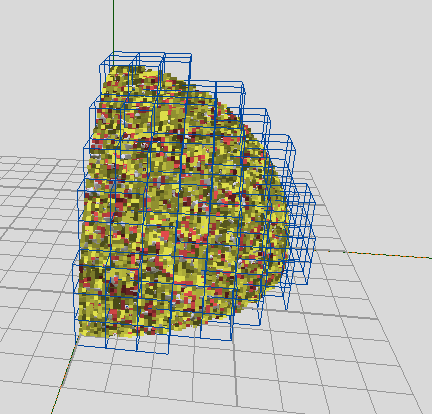}
    \caption{Random Raycasting}
    \label{fig:random_raycast}
  \end{subfigure}
  \vspace*{-0.5cm}
  \end{table}

We first compare different versions of our method with each other and against baselines from other works. We use random insertions of pointclouds with different sizes as a baseline setting. All experimental results are averaged over 5 runs. The results are presented in Table~\ref{tab:runtime}. Even the basis variant using OpenVDB is able to outperform OctoMap by a factor of 2, in multiple settings this advantage grows up to a factor of 30. Interestingly, the parallel integration scheme \textit{VDB-PAR} achieves almost linear speedup for small ray lengths. Figure~\ref{fig:eval_threads} gives further insight into that evaluation. The reason for the performance drop for higher ray lengths is the increasing amount of time required for the \textit{merge} operations as insertion performance stays constant for increasing submap sizes while the merge workload grows cubic. This insight can be derived from Figure~\ref{fig:eval_length}, where the different steps of an integration procedure are measured. Consequently, the best parallelization speedup can be achieved in low-range settings with a lot of points to be integrated. This exactly matches the domain of indoor mapping scenarios using high-resolution solid-state LiDARs as we will show in the next sectione A detailed evaluation of different ray lengths is given in Figure~\ref{fig:eval_full}. OctoMap and the VDB-EDT approach from~\cite{vdb-edt} are outperformed in low range (below 30m) scenarios by almost a magnitude.

The \textit{bundling} optimization VDB-BUN guarantees to save runtime, as the first ray to a specific voxel is skipped in every case. This approximately halves the runtime over all settings. The \textit{subsampling} optimization VDB-SUB on the other hand only applies when many points reside in a small volume. Consequently it saves the most runtime in a setting with many points in a \textit{structured} environment, whereas it is not faster or even slower for different settings.

\begin{table}[bt]
  \caption{Benchmarks on the \textit{cow-and-lady} dataset. \textit{\#Points} denote the amount of processed points over all frames. This varies due to different processing speeds and different temporal alignments bewteen poses and pointcloud. \textit{Time} measures the total integration time and \textit{\#Occupied Voxels} counts the number of occupied voxels after the integration procedure.}
  \label{tab:datasets}
  \renewcommand{\arraystretch}{1.1}
  \centering
  \begin{tabular}{@{}lrrr}
    \toprule
    Name  & \#Points & \makecell{\#Occupied\\ Voxels} & \makecell{Time per \\ frame [ms]} \\ \midrule
    OctoMap~\cite{octomap} &  $0.463 \cdot 10^{9}$ & $0.332 \cdot 10^{6}$ & 388 \\
    VDB-EDT~\cite{vdb-edt} & $0.557 \cdot 10^{9}$ & $0.567 \cdot 10^{6}$ & 357 \\
    VDB-MAP~\cite{vdb_fzi} & $0.518 \cdot 10^{9}$ & $0.536 \cdot 10^{6}$ & 263 \\
    VDB-FMAP & $0.513 \cdot 10^{9}$ & $0.523 \cdot 10^{6}$ & 282 \\
    VDB-BUN & $0.522 \cdot 10^{9}$ & $0.316 \cdot 10^{6}$ & 94 \\
    VDB-SUB & $0.521 \cdot 10^{9}$ & $0.510 \cdot 10^{6}$ & 170 \\
    VDB-PAR & $0.526 \cdot 10^{9}$ & $0.530 \cdot 10^{6}$ & $\mathbf{47}$ \\ \bottomrule
  \end{tabular}
\end{table}
\begin{figure}[tb]
  \centering
  \begin{subfigure}[b]{0.23\textwidth}
    \centering
    \includegraphics[width=\textwidth]{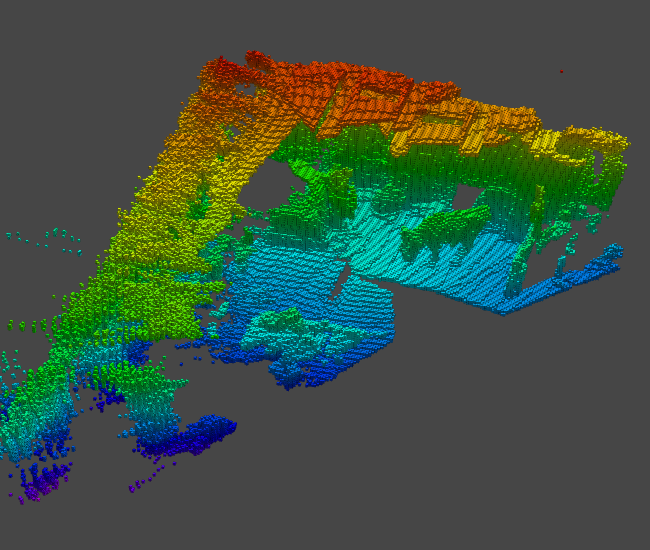}
    \caption{VDB-FMAP \((100\%)\)}
    \label{fig:gaz_raycasting}
  \end{subfigure}
  \hfill
  \begin{subfigure}[b]{0.23\textwidth}
    \centering
    \includegraphics[width=\textwidth]{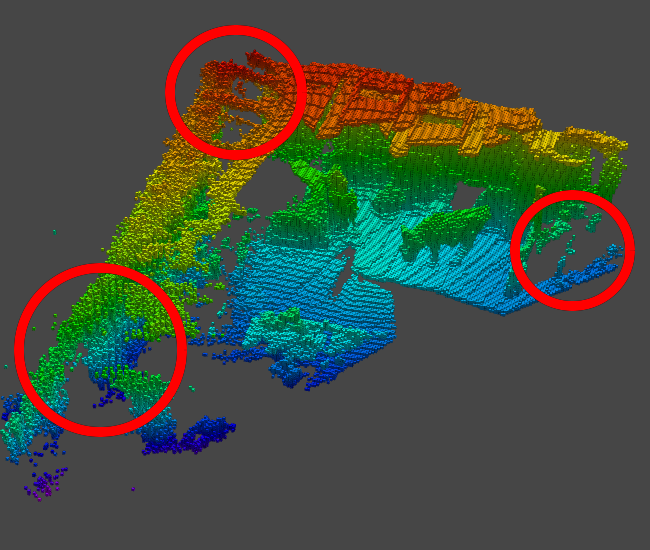}
    \caption{VDB-BUN \((23.0\%)\)}
    \label{fig:gaz_bundle}
  \end{subfigure}\\[0.3cm]
  \begin{subfigure}[b]{0.23\textwidth}
    \centering
    \includegraphics[width=\textwidth]{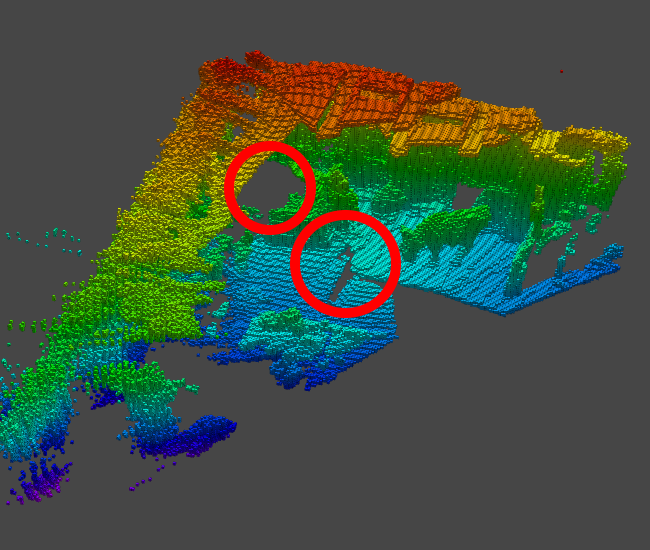}
    \caption{VDB-SUB \((39.7\%)\)}
    \label{fig:gaz_subsample}
  \end{subfigure}
  \hfill
  \begin{subfigure}[b]{0.23\textwidth}
    \centering
    \includegraphics[width=\textwidth]{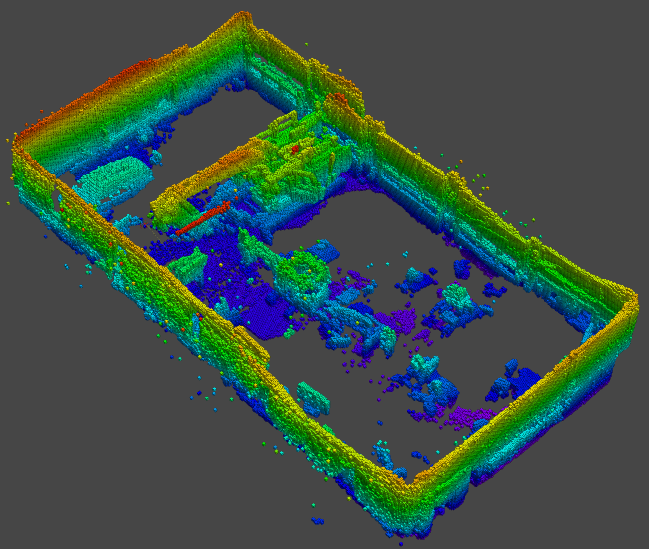}
    \caption{Indoor lab map \((100\%)\)}
    \label{fig:sim_hall}
  \end{subfigure}
  \caption{Qualitative visualization of different mapping optimization effects on the \textit{cow-and-lady} dataset in (a)-(c) as well as the lab environment captured by an indoor UAV (cf.~\ref{fig:drone}) in (d). The percentage denotes how many of the original points are casted as rays using the respective optimization.}
  \label{fig:mapping_sim}
\end{figure}

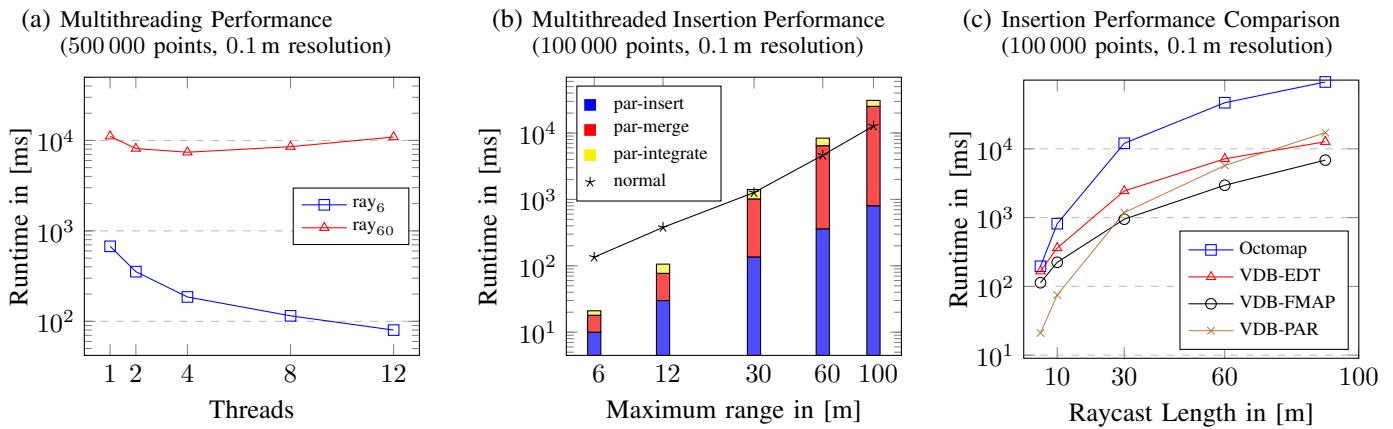
\begin{figure*}[bt]
  \captionsetup{position=top, font=normal, labelfont=normal}
  \centering
  \begin{subfigure}[t]{0.31\textwidth}
    \captionsetup{format=hang, margin={0.3cm,0.1cm}}
    \centering
    \caption{Multithreading Performance \\ (\(500\,000\) points, \(0.1\,\)m resolution)}
    \begin{tikzpicture}
      \begin{axis}[
          scale=0.65,
          xlabel={Threads},
          ylabel={Runtime in [ms]},
          ylabel shift = -5 pt,
          xmin=0, xmax=13,
          ymin=0, ymax=50000,
          xtick={1,2,4,8,12},
          ymode=log,
          log basis y={10},
          legend style={at={(0.97,0.5)},anchor=east},
          legend cell align={left},
          ymajorgrids=true,
          grid style=dashed,
          ]
          \addplot[
          color=blue,
          mark=square,
          ]
          coordinates {
              (1,675)(2,354)(4,186)(8,115)(12,80)
            };
          \legend{\scriptsize{$\text{ray}_{6}$}}
          \addplot[
          color=red,
          mark=triangle,
          ]
          coordinates {
              (1,11121)(2,8134)(4,7405)(8,8530)(12,10904)
            };
          \legend{\scriptsize{$\text{ray}_{6}$}, \scriptsize{$\text{ray}_{60}$}}
        \end{axis}
      \end{tikzpicture}
      \label{fig:eval_threads}
  \end{subfigure}\hfill
  \begin{subfigure}[t]{0.31\textwidth}
    \captionsetup{format=hang, margin={0.3cm,0.1cm}}
    \centering
    \caption{Multithreaded Insertion Performance  \\ (\(100\,000\) points, \(0.1\,\)m resolution)}
    \begin{tikzpicture}[
    every axis/.style={ % add these settings to all the axis environments in the tikzpicture
        scale=0.65,
        xmode=log,
        ymode=log,
        xtick={6, 12, 30, 60, 100},
        xlabel={Maximum range in [m]},
        ylabel={Runtime in [ms]},
        y domain=0:60000
        ytick={100, 1000, 10000, 50000},
        bar width=5pt,
        log x ticks with fixed point,
        % log y ticks with fixed point,
    },]
    \begin{axis}[legend pos=north west, legend cell align={left}, legend style={row sep=0pt, column sep=5pt}]
      \addlegendimage{only marks, mark=square*, blue};
      \addlegendimage{only marks, mark=square*, red};
      \addlegendimage{only marks, mark=square*, yellow};
      \addlegendimage{only marks, mark=star};
      \addlegendentry{\scriptsize{par-insert}}
      \addplot [fill=blue!70, bar shift=0pt, ybar stacked] coordinates {(6,10) (12,30) (30,136) (60,361) (100, 803)};
      \addlegendentry{\scriptsize{par-merge}}
      \addplot [fill=red!70, bar shift=0pt, ybar stacked] coordinates {(6,8) (12,47) (30,881) (60,6085) (100,24405)};
      \addlegendentry{\scriptsize{par-integrate}}
      \addplot [fill=yellow!70, bar shift=0pt, ybar stacked] coordinates {(6,3) (12,29) (30,393) (60,1943) (100,5801)};
      \addlegendentry{\scriptsize{normal}}
      \addplot +[] coordinates {(6,135) (12,379) (30,1269) (60,4631) (100, 12780)};
    \end{axis}
    \end{tikzpicture}
  \label{fig:eval_length}%
  \end{subfigure}\hfill
  \begin{subfigure}[t]{0.31\textwidth}
    \captionsetup{format=hang, margin={0.3cm,0.1cm}}
    \centering
    \caption{Insertion Performance Comparison  \\ (\(100\,000\) points, \(0.1\,\)m resolution)}
    \begin{tikzpicture}
      \begin{axis}[
        scale=0.65,
        xlabel={Raycast Length in [m]},
        ylabel={Runtime in [ms]},
        ylabel shift = -5 pt,
        xmin=0, xmax=100,
        ymin=0, ymax=100000,
        xtick={10, 30, 60, 100},
        ytick={10, 100, 1000, 10000},
        ymode=log,
        log basis y={10},
        legend pos=south east,
        legend cell align={left},
        ymajorgrids=true,
        grid style=dashed,
        ]
        \addplot[
        color=blue,
        mark=square,
        ]
        coordinates {
          (5,196)(10,816)(30,12044)(60,46867)(90,93942)
        };
        \addplot[
        color=red,
        mark=triangle,
        ]
        coordinates {
          (5,166)(10,362)(30,2440)(60,7167)(90,12692)
        };
        \addplot[
        color=black,
        mark=o,
        ]
        coordinates {
          (5,113)(10,224)(30,948)(60,2957)(90,6857)
        };
        \addplot[
        color=brown,
        mark=x,
        ]
        coordinates {
          (5,21)(10,75)(30,1183)(60,5726)(90,17212)
        };
        \legend{\scriptsize{Octomap}, \scriptsize{VDB-EDT}, \scriptsize{VDB-FMAP}, \scriptsize{VDB-PAR}}
      \end{axis}
    \end{tikzpicture}
    \label{fig:eval_full}
  \end{subfigure}
  \captionsetup{position=bottom, font=normal, labelfont=normal}
  \caption{Evaluation and comparison of map integration procedures. (a) and (b) examines the influence of multithreading on the integration scheme, while (c) compares different ray distance limits with other approaches. Note the log-scale at the runtime.}
  \label{fig:map_integration_eval}
\end{figure*}

\subsection{Benchmark and Real Datasets}
We evaluate the presented methods on the indoor \textit{cow-and-lady} dataset released as part of VoxBlox~\cite{voxblox}. It consists of 2831 depth frames captured by a Microsoft Kinect I as well as corresponding pose data. The results in Table~\ref{tab:datasets} show that VDB based methods outperform previous methods in terms of simple integration performance. For the \textit{bundling} optimization VDB-BUN, the resulting amount of occupied voxels is only half compared to the other variants. This indicates that the map quality is reduced. On the other hand, the \textit{subsampling} strategy VDB-SUB only comes along with a negliglible reduction of occupied voxels. This shows that mostly redundant rays are omitted in the integration procedure while it is almost able to halve the runtime. Again, the parallel version outperforms all other versions without any reduction in accuracy.

Figure~\ref{fig:mapping_sim} examplarily shows an excerpt of the \textit{cow-and-lady} dataset after 500 frames for different integration strategies. There are only few details (marked in red) of quality loss for saving more than \(60\%\) of the integration steps using the subsampling optimization and almost 80\% using the bundled integration. Figures~\ref{fig:mapping_iosb} and~\ref{fig:sim_hall} show that our mapping procedure is capable of producing high quality maps of outdoor and indoor environments using a legged robot or a drone, respectively. Even without preprocessing the poses using a factor graph~\cite{Dellaert17fnt}, the resulting map quality is satisfactory, if the pose drift is not too large (cf.~\ref{fig:sim_hall}).

\section{Conclusion and Discussion}\label{sec:discussion}

We presented an efficient map building approach for mobile robots, especially suitable for agile robots in dynamic environments. The experimental evaluation demonstrates the effectiveness of the approach particularly in indoor environments with low sensor ranges, where the approach outperforms current solutions. The parallel fusion of different maps allows not only fast but also flexible integration of new sensor data into the map. This leaves room for additional improvements such as an extension with dynamic resolution adaption. Maintaining a real-time distance map is also a valuable extension, which could be implemented efficiently using the VDB datastructure. In order to reduce global inconsistencies coming from drifts in the localization, one could couple the VDB representation with a factor graph backend. This has been tested in a prototyped fashion as demonstrated in Figure~\ref{fig:mapping_iosb} but can potentially be improved by a tighter coupling.

\bibliographystyle{IEEEtran}
\bibliography{IEEEabrv, literature}

\begin{thebibliography}{10}
\providecommand{\url}[1]{#1}
\csname url@samestyle\endcsname
\providecommand{\newblock}{\relax}
\providecommand{\bibinfo}[2]{#2}
\providecommand{\BIBentrySTDinterwordspacing}{\spaceskip=0pt\relax}
\providecommand{\BIBentryALTinterwordstretchfactor}{4}
\providecommand{\BIBentryALTinterwordspacing}{\spaceskip=\fontdimen2\font plus
\BIBentryALTinterwordstretchfactor\fontdimen3\font minus
  \fontdimen4\font\relax}
\providecommand{\BIBforeignlanguage}[2]{{%
\expandafter\ifx\csname l@#1\endcsname\relax
\typeout{** WARNING: IEEEtran.bst: No hyphenation pattern has been}%
\typeout{** loaded for the language `#1'. Using the pattern for}%
\typeout{** the default language instead.}%
\else
\language=\csname l@#1\endcsname
\fi
#2}}
\providecommand{\BIBdecl}{\relax}
\BIBdecl

\bibitem{herbertmapping}
M.~Herbert, C.~Caillas, E.~Krotkov, I.~Kweon, and T.~Kanade, ``{Terrain Mapping
  for a Roving Planetary Explorer},'' in \emph{Proceedings, 1989 International
  Conference on Robotics and Automation}, 1989, pp. 997--1002.

\bibitem{iosb_mapping}
T.~Emter and J.~Petereit, ``{3D} {SLAM} {W}ith {S}can {M}atching and {F}actor
  {G}raph {O}ptimization,'' in \emph{ISR 2018; 50th International Symposium on
  Robotics}, 2018.

\bibitem{octomap}
A.~Hornung, K.~M. Wurm, M.~Bennewitz, C.~Stachniss, and W.~Burgard, ``{OctoMap:
  An Efficient Probabilistic 3D Mapping Framework Based on Octrees},''
  \emph{Autonomous Robots}, 2013.

\bibitem{voxblox}
H.~Oleynikova, Z.~Taylor, M.~Fehr, R.~Siegwart, and J.~Nieto, ``{Voxblox:}
  {I}ncremental {3D} {E}uclidean {S}igned {D}istance {F}ields for {On-Board}
  {MAV} {P}lanning,'' in \emph{IEEE/RSJ International Conference on Intelligent
  Robots and Systems (IROS)}, 2017.

\bibitem{sdf}
H.~Oleynikova, A.~Millane, Z.~Taylor, E.~Galceran, J.~Nieto, and R.~Siegwart,
  ``{Signed Distance Fields: A Natural Representation for Both Mapping and
  Planning},'' 2016.

\bibitem{vdb_fzi}
M.~Grosse~Besselmann, L.~Puck, L.~Steffen, A.~Roennau, and R.~Dillmann,
  ``{VDB-Mapping: A High Resolution and Real-Time Capable 3D Mapping Framework
  for Versatile Mobile Robots},'' 2021.

\bibitem{vdb-edt}
D.~Zhu, C.~Wang, W.~Wang, R.~Garg, S.~A. Scherer, and M.~Q. Meng, ``{VDB-EDT:
  An Efficient Euclidean Distance Transform Algorithm Based on VDB Data
  Structure},'' \emph{CoRR}, vol. abs/2105.04419, 2021.

\bibitem{spatiotemporal}
S.~Macenski, D.~Tsai, and M.~Feinberg, ``Spatio-temporal voxel layer: A view on
  robot perception for the dynamic world,'' \emph{International Journal of
  Advanced Robotic Systems}, vol.~17, no.~2, 2020.

\bibitem{ros2}
S.~Macenski, T.~Foote, B.~Gerkey, C.~Lalancette, and W.~Woodall, ``Robot
  operating system 2: Design, architecture, and uses in the wild,''
  \emph{Science Robotics}, vol.~7, no.~66, p. eabm6074, 2022.

\bibitem{vdb}
K.~Museth, J.~Lait, J.~Johanson, J.~Budsberg, R.~Henderson, M.~Alden, P.~Cucka,
  D.~Hill, and A.~Pearce, ``{OpenVDB: An Open-Source Data Structure and Toolkit
  for High-Resolution Volumes},'' in \emph{ACM SIGGRAPH 2013 Courses}, ser.
  SIGGRAPH '13.\hskip 1em plus 0.5em minus 0.4em\relax New York, NY, USA:
  Association for Computing Machinery, 2013.

\bibitem{bplustree}
R.~Bayer and E.~McCreight, ``{Organization and Maintenance of Large Ordered
  Indices},'' in \emph{Proceedings of the 1970 ACM SIGFIDET (Now SIGMOD)
  Workshop on Data Description, Access and Control}, ser. SIGFIDET '70, 1970,
  p. 107–141.

\bibitem{logodds}
H.~Moravec and A.~Elfes, ``{High Resolution Maps from Wide Angle Sonar},'' in
  \emph{Proceedings. 1985 IEEE International Conference on Robotics and
  Automation}, vol.~2, 1985, pp. 116--121.

\bibitem{Dellaert17fnt}
F.~Dellaert and M.~Kaess, \emph{{Factor Graphs for Robot Perception}}, 2017.

\end{thebibliography}

\end{document}